\documentclass[10pt]{article}
\usepackage[dvips]{graphicx}
\usepackage{amsmath}
\usepackage{amsfonts}
\usepackage{amscd}
\usepackage{epsf}

\newtheorem{pseudo}{Algorithm}

\begin{document}

\title{Selection in Scale-Free Small World}

\author{Zs.~Palotai$^1$, Cs.~Farkas$^2$, A.~L{\H
o}rincz$^1$\thanks{Corresponding author. email:
lorincz@inf.elte.hu}}

\date{}

\maketitle

\small{ \noindent$^1$ Eotvos Lorand University, Department of
Information Systems,
Pazmany Peter setany 1/c, Budapest, H1117, Hungary\\
$^2$ University of South Carolina, Department of Computer Sciences
and Engineering, Columbia, SC 29208, USA}

\begin{abstract}
In this paper we compare the performance characteristics of our
selection based learning algorithm for Web crawlers with the
characteristics of the reinforcement learning algorithm. The task
of the crawlers is to find new information on the Web. The
selection algorithm, called weblog update, modifies the starting
URL lists of our crawlers based on the found URLs containing new
information. The reinforcement learning algorithm modifies the URL
orderings of the crawlers based on the received reinforcements for
submitted documents. We performed simulations based on data
collected from the Web. The collected portion of the Web is
typical and exhibits scale-free small world (SFSW) structure. We
have found that on this SFSW, the weblog update algorithm performs
better than the reinforcement learning algorithm. It finds the new
information faster than the reinforcement learning algorithm and
has better new information/all submitted documents ratio. We
believe that the advantages of the selection algorithm over
reinforcement learning algorithm is due to the small world
property of the Web.
\end{abstract}

% ------------------------------------------------------------------------
\section{Introduction}\label{s:intro}

The largest source of information today is the World Wide Web. The
estimated number of documents nears 10 billion. Similarly, the
number of documents changing on a daily basis is also enormous.
The ever-increasing growth of the Web presents a considerable
challenge in finding novel information on the Web.

In addition, properties of the Web, like scale-free small world (SFSW) structure
\cite{barabasi00scalefree,Kleinberg01structure} may create additional challenges. For example the direct consequence of
the scale-free small world property is that there are numerous URLs or sets of interlinked URLs, which have a large
number of incoming links. Intelligent web crawlers can be easily trapped at the neighborhood of such junctions as it
has been shown previously \cite{kokai02learning,Lorincz02intelligent}.

%Because of this we have developed a novel artificial life (A-life) method with intelligent individuals, crawlers, to
%detect new information on a news web site. We define A-life as a population of individuals having both wired structural
%properties, and structural properties which may undergo continuous changes, i.e., adaptation. Our algorithms are based
%on methods developed in different areas of artificial intelligence, such as evolutionary computing, artificial neural
%networks and reinforcement learning.

We have developed a novel artificial life (A-life) method with
intelligent individuals, crawlers, to detect new information on a
news Web site. We define A-life as a population of individuals
having both static structural properties, and structural
properties which may undergo continuous changes, i.e., adaptation.
Our algorithms are based on methods developed for different areas
of artificial intelligence, such as evolutionary computing,
artificial neural networks and reinforcement learning. All efforts
were made to keep the applied algorithms as simple as possible
subject to the constraints of the internet search.

Evolutionary computing deals with properties that may be modified
during the creation of new individuals, called 'multiplication'.
Descendants may exhibit variations of population, and differ in
performance from the others. Individuals may also terminate.
Multiplication and selection is subject to the fitness of
individuals, where  fitness is typically defined by the modeler.
For a recent review on evolutionary computing, see
\cite{eiben03introduction}. For reviews on related evolutionary
theories and the dynamics of self-modifying systems see
\cite{Fryxell98individual,Clarck00dynamic} and
\cite{kampis91selfmodifying,Csanyi89evolutionary}, respectively.
Similar concepts have been studied in other evolutionary systems
where organisms compete for space and resources and cooperate
through direct interaction (see, e.g., \cite{taylor02mutualism}
and references therein.)

Selection, however, is a very slow process and individual adaptation may be necessary in environments subject to quick
changes. The typical form of adaptive learning is the connectionist architecture, such as artificial neural networks.
Multilayer perceptrons (MLPs), which are universal function approximators have been used widely in diverse
applications. Evolutionary selection of adapting MLPs has been in the focus of extensive research
\cite{yao93review,yao99evolving}.

In a typical reinforcement learning (RL) problem the learning process \cite{Sutton98Reinforcement} is motivated by the
expected value of long-term cumulated profit. A well-known example of reinforcement learning is the TD-Gammon program
of Tesauro \cite{tesauro95temporal}. The author applied MLP function approximators for value estimation. Reinforcement
learning has also been used in concurrent multi-robot learning, where robots had to learn to forage together via direct
interaction \cite{mataric97reinforcement}. Evolutionary learning has been used within the framework of reinforcement
learning to improve decision making, i.e., the state-action mapping called policy
\cite{stafylopatis98autonomous,moriarty99evolutionary,tuyls03extended,kondo04reinforcement}.

In this paper we present a selection based algorithm and compare it to the well-known reinforcement learning algorithm
in terms of their efficiency and behavior. In our problem, fitness is not determined by us, but fitness is implicit.
Fitness is jointly determined by the ever changing external world and by the competing individuals together. Selection
and multiplication of individuals are based on their fitness value. Communication and competition among our crawlers
are indirect. Only the first submitter of a document may receive positive reinforcement. Our work is different from
other studies using combinations of genetic, evolutionary, function approximation, and reinforcement learning
algorithms, in that i) it does not require explicit fitness function, ii) we do not have control over the environment,
iii) collaborating individuals use value estimation under `evolutionary pressure', and iv) individuals work without
direct interaction with each other.

We performed realistic simulations based on data collected during
an 18 days long crawl on the Web. We have found that our selection
based weblog update algorithm performs better in scale-free small
world environment than the RL algorithm, eventhough the
reinforcement learning algorithm has been shown to be efficient in
finding relevant information
\cite{Lorincz02intelligent,rennie99using}. We explain our results
based on the different behaviors of the algorithms. That is, the
weblog update algorithm finds the good relevant document sources
and remains at these regions until better places are found by
chance. Individuals using this selection algorithm are able to
quickly collect the new relevant documents from the already known
places because they monitor these places continuously. The
reinforcement learning algorithm explores new territories for
relevant documents and if it finds a good place then it collects
the existing relevant documents from there. The continuous
exploration of RL causes that it finds relevant documents slower
than the weblog update algorithm. Also, crawlers using weblog
update algorithm submit more different documents than crawlers
using the RL algorithm. Therefore there are more relevant new
information among documents submitted by former than latter
crawlers.

The paper is organized as follows. In Section \ref{s:related} we review recent works in the field of Web crawling. Then
we describe our algorithms and the forager architecture in Section \ref{s:architecture}. After that in Section
\ref{s:experiments} we present our experiment on the Web and the conducted simulations with the results. In Section
\ref{s:discussion} we discuss our results on the found different behaviors of the selection and reinforcement learning
algorithms. Section \ref{s:conclusion} concludes our paper.

\section{Related work}\label{s:related}

Our work concerns a realistic Web environment and search
algorithms over this environment. We compare
selective/evolutionary and reinforcement learning methods. It
seems to us that such studies should be conducted in ever
changing, buzzling, wabbling environments, which justifies our
choice of the environment. We shall review several of the known
search tools including those
\cite{kokai02learning,Lorincz02intelligent} that our work is based
upon. Readers familiar with search tools utilized on the Web may
wish to skip this section.

There are three main problems that have been studied in the context of crawlers. Rungsawang et al.
\cite{rungsawang04learnable} and references therein and Menczer \cite{menczer03complementing} studied the topic
specific crawlers. Risvik et al. \cite{risvik02search} and references therein address research issues related to the
exponential growth of the Web. Cho and Gracia-Molina \cite{cho03effective}, Menczer \cite{menczer03complementing} and
Edwards et. al \cite{edwards01adaptive} and references therein studies the problem of different refresh rates of URLs
(possibly as high as hourly or as low as yearly).

Rungsawang and Angkawattanawit \cite{rungsawang04learnable} provide an introduction to and a broad overview of topic
specific crawlers (see citations in the paper). They propose to learn starting URLs, topic keywords and URL ordering
through consecutive crawling attempts. They show that the learning of starting URLs and the use of consecutive crawling
attempts can increase the efficiency of the crawlers. The used heuristic is similar to the weblog algorithm
\cite{Gabor04value}, which also finds good starting URLs and periodically restarts the crawling from the newly learned
ones. The main limitation of this work is that it is incapable of addressing the freshness (i.e., modification) of
already visited Web pages.

Menczer \cite{menczer03complementing} describes some disadvantages of current Web search engines on the dynamic Web,
e.g., the low ratio of fresh or relevant documents. He proposes to complement the search engines with intelligent
crawlers, or web mining agents to overcome those disadvantages. Search engines take static snapshots of the Web with
relatively large time intervals between two snapshots. Intelligent web mining agents are different: they can find
online the required recent information and may evolve intelligent behavior by exploiting the Web linkage and textual
information.

He introduces the InfoSpider architecture that uses genetic algorithm and reinforcement learning, also describes the
MySpider implementation of it. Menczer discusses the difficulties of evaluating online query driven crawler agents. The
main problem is that the whole set of relevant documents for any given query are unknown, only a subset of the relevant
documents may be known. To solve this problem he introduces two new metrics that estimate the real recall and precision
based on an available subset of the relevant documents. With these metrics search engine and online crawler
performances can be compared. Starting the MySpider agent from the 100 top pages of AltaVista the agent's precision is
better than AltaVista's precision even during the first few steps of the agent.

The fact that the MySpider agent finds relevant pages in the first few steps may make it deployable on users'
computers. Some problems may arise from this kind of agent usage. First of all there are security issues, like which
files or information sources are allowed to read and write for the agent. The run time of the agents should be
controlled carefully because there can be many users (Google answered more than 100 million searches per day in
January-February 2001) using these agents, thus creating huge traffic overhead on the Internet.

Our weblog algorithm uses local selection for finding good starting URLs for searches, thus not depending on any search
engines. Dependence on a search engine can be a suffer limitation of most existing search agents, like MySpiders. Note
however, that it is an easy matter to combine the present algorithm with URLs offered by search engines. Also our
algorithm should not run on individual users's computers. Rather it should run for different topics near to the source
of the documents in the given topic -- e.g., may run at the actual site where relevant information is stored.

Risvik and Michelsen \cite{risvik02search} mention that because of the exponential growth of the Web there is an ever
increasing need for more intelligent, (topic-)specific algorithms for crawling, like focused crawling and document
classification. With these algorithms crawlers and search engines can operate more efficiently in a topically limited
document space. The authors also state that in such vertical regions the dynamics of the Web pages is more homogenous.

They overview different dimensions of web dynamics and show the arising problems in a search engine model. They show
that the problem of rapid growth of Web and frequent document updates creates new challenges for developing more and
more efficient Web search engines. The authors define a reference search engine model having three main components: (1)
crawler, (2) indexer, (3) searcher. The main part of the paper focuses on the problems that crawlers need to overcome
on the dynamic Web. As a possible solution the authors propose a heterogenous crawling architecture. They also present
an extensible indexer and searcher architecture. The crawling architecture has a central distributor that knows which
crawler has to crawl which part of the web. Special crawlers with low storage and high processing capacity are
dedicated to web regions where content changes rapidly (like news sites). These crawlers maintain up-to-date
information on these rapidly changing Web pages.

The main limitation of their crawling architecture is that they must divide the web to be crawled into distinct
portions manually before the crawling starts. A weblog like distributed algorithm -- as suggested here -- my be used in
that architecture to overcome this limitation.

Cho and Garcia-Molina \cite{cho03effective} define mathematically the freshness and age of documents of search engines.
They propose the Poisson process as a model for page refreshment. The authors also propose various refresh policies and
study their effectiveness both theoretically and on real data. They present the optimal refresh policies for their
freshness and age metrics under the Poisson page refresh model. The authors show that these policies are superior to
others on real data, too.

They collected about 720000 documents from 270 sites. Although
they show that in their database more than 20 percent of the
documents are changed each day, they disclosed these documents
from their studies. Their crawler visited the documents once each
day for 5 months, thus can not measure the exact change rate of
those documents. While in our work we definitely concentrate on
these frequently changing documents.

The proposed refresh policies require good estimation of the
refresh rate for each document. The estimation influences the
revisit frequency while the revisit frequency influences the
estimation. Our algorithm does not need explicit frequency
estimations. The more valuable URLs (e.g., more frequently
changing) will be visited more often and if a crawler does not
find valuable information around an URL being in it's weblog then
that URL finally will fall out from the weblog of the crawler.
However frequency estimations and refresh policies can be easily
integrated into the weblog algorithm selecting the starting URL
from the weblog according to the refresh policy and weighting each
URL in the weblog according to their change frequency estimations.

Menczer \cite{menczer03complementing} also introduces a recency metric which is 1 if all of the documents are recent
(i.e., not changed after the last download) and goes to 0 as downloaded documents are getting more and more obsolete.
Trivially immediately after a few minutes run of an online crawler the value of this metric will be 1, while the value
for the search engine will be lower.

Edwards et al. \cite{edwards01adaptive} present a mathematical crawler model in which the number of obsolete pages can
be minimized with a nonlinear equation system. They solved the nonlinear equations with different parameter settings on
realistic model data. Their model uses different buckets for documents having different change rates therefore does not
need any theoretical model about the change rate of pages. The main limitations of this work are the following:

\begin{itemize}
  \item by solving the nonlinear
equations the content of web pages can not be taken into
consideration. The model can not be extended easily to
(topic-)specific crawlers, which would be highly advantageous on
the exponentially growing web \cite{rungsawang04learnable},
\cite{risvik02search}, \cite{menczer03complementing}.
  \item the rapidly changing documents (like on news sites) are
  not considered to be in any bucket, therefore increasingly
  important parts of the web are disclosed from the searches.
\end{itemize}

However the main conclusion of the paper is that there may exist some efficient strategy for incremental crawlers for
reducing the number of obsolete pages without the need for any theoretical model about the change rate of pages.

%\section{Framework}\label{s:framework}

%Scale-free Small World fast changing environment.

\section{Forager architecture}\label{s:architecture}

There are two different kinds of agents: the foragers and the reinforcing agent (RA). The fleet of foragers crawl the
web and send the URLs of the selected documents to the reinforcing agent. The RA determines which forager should work
for the RA and how long a forager should work. The RA sends reinforcements to the foragers based on the received URLs.

We employ a fleet of foragers to study the competition among individual foragers. The fleet of foragers allows to
distribute the load of the searching task among different computers. A forager has simple, limited capabilities, like
limited number of starting URLs and a simple, content based URL ordering. The foragers compete with each other for
finding the most relevant documents. In this way they efficiently and quickly collect new relevant documents without
direct interaction.

At first the basic algorithms are presented. After that the reinforcing agent and the foragers are detailed.

\subsection{Algorithms}\label{ss:algorithms}

\subsubsection{Weblog algorithm and starting URL
selection}\label{sss:weblog}

A forager periodically restarts from a URL randomly selected from the list of starting URLs. The sequence of visited
URLs between two restarts forms a path. The starting URL list is formed from the $START\_SIZE=10$ first URLs of the
weblog. In the weblog there are $WEBLOG\_SIZE=100$ URLs with their associated weblog values in descending order. The
weblog value of a URL estimates the expected sum of rewards during a path after visiting that URL. The weblog update
algorithm modifies the weblog before a new path is started (Algorithm \ref{t:weblog_pseudo}). The weblog value of a URL
already in the weblog is modified toward the sum of rewards in the remaining part of the path after that URL. A new URL
has the value of actual sum of rewards in the remaining part of the path. If a URL has a high weblog value it means
that around that URL there are many relevant documents. Therefore it may worth it to start a search from that URL.

\begin{table}[htb]\hrule\vskip1pt\hrule\vskip2mm
\begin{pseudo}\label{t:weblog_pseudo}\normalfont\textbf{Weblog Update}. $\beta$ was set to 0.3
  \vskip1mm \hrule %\vskip1pt \hrule
  \begin{tabbing}
  xxx \= xx \= xx \= xx \= xx \= xx \= xx \= xx \= xx \= \kill
  \verb"input"\\
  \> $visitedURLs\leftarrow$ the steps of the given
  path \\
  \> $values\leftarrow$ the sum of rewards for
  each step in the given path \\
  \verb"output"\\
  \> starting URL list\\
  \verb"method"\\
  \> $cumValues\leftarrow$ cumulated sum of
  $values$ in reverse order\\
  \> $newURLs\leftarrow visitedURLs$ not having value in
  $weblog$\\
  \> $revisitedURLs \leftarrow visitedURLs$ having value
  in $weblog$\\
  \> \verb"for each" $URL\,\in\,newURLs$\\
  \> \> $weblog(URL)\leftarrow cumValues(URL)$ \\
  \> \verb"endfor"\\
  \> \verb"for each" $URL\,\in\,revisitedURLs$\\
  \> \> $weblog(URL)\leftarrow (1-\beta)\,weblog(URL)\,+$\\
  \> \> \> $\beta\,cumValues(URL)$ \\
  \> \verb"endfor"\\
  \> $weblog\leftarrow$ descending order of values in $weblog$\\
  \> $weblog\leftarrow$ truncate $weblog$ after the $WEBLOG\_SIZE^{th}$\\
  \> \> element\\
  \> starting URL list $\leftarrow$ first $START\_SIZE$ elements of
  $weblog$
  \end{tabbing}
  \hrule \vskip1pt \hrule
%\end{table}
\end{pseudo}
\end{table}

%Without the weblog algorithm the weblog and thus the starting URL list remains the same throughout the searches.

Without the weblog algorithm the weblog and thus the starting URL
list remains the same throughout the searches. The weblog
algorithm is a very simple version of evolutionary algorithms.
Here, evolution may occur at two different levels: the list of
URLs of the forager is evolving by the reordering of the weblog.
Also, a forager may multiply, and its weblog, or part of it may
spread through inheritance. This way, the weblog algorithm
incorporates most basic features of evolutionary algorithms. This
simple form shall be satisfactory to demonstrate our statements.

\subsubsection{Reinforcement Learning and URL
ordering}\label{sss:rl}

A forager can modify its URL ordering based on the received reinforcements of the sent URLs. The (immediate) profit is
the difference of received rewards and penalties at any given step. Immediate profit is a myopic characterization of a
step to a URL. Foragers have an adaptive continuous value estimator and follow the \textit{policy} that maximizes the
expected long term cumulated profit (LTP) instead of the immediate profit. Such estimators can be easily realized in
neural systems \cite{Sutton98Reinforcement,szita03kalman,schultz00multiple}. Policy and profit estimation are
interlinked concepts: profit estimation determines the policy, whereas policy influences choices and, in turn, the
expected LTP. (For a review, see \cite{Sutton98Reinforcement}.) Here, choices are based on the greedy LTP policy: The
forager visits the URL, which belongs to the \textit{frontier} (the list of linked but not yet visited URLs, see later)
and has the highest estimated LTP.

In the particular simulation each forager has a $k(=50)$ dimensional probabilistic term-frequency inverse
document-frequency (PrTFIDF) text classifier \cite{joachims97probabilistic}, generated on a previously downloaded
portion of the Geocities database. Fifty clusters were created by Boley's clustering algorithm \cite{Boley98principal}
from the downloaded documents. The PrTFIDF classifiers were trained on these clusters plus an additional one, the
$(k+1)^{th}$, representing general texts from the internet. The PrTFIDF outputs were non-linearly mapped to the
interval [-1,+1] by a hyperbolic-tangent function. The classifier was applied to reduce the texts to a small
dimensional representation. The output vector of the classifier for the page of URL $A$ is
$\mathbf{state(A)}=(state(A)_1,\ldots , state(A)_k)$. (The $(k+1)^{th}$ output was dismissed.) This output vector is
stored for each URL (Algorithm \ref{t:pageinfo_URLordering_pseudo}).

\begin{table}[htb]\hrule\vskip1pt\hrule\vskip2mm
\begin{pseudo}\label{t:pageinfo_URLordering_pseudo}\normalfont\textbf{Page Information Storage}
  \vskip1mm \hrule %\vskip1pt \hrule
  \begin{tabbing}
  xxx \= xx \= xx \= xx \= xx \= xx \= xx \= xx \= xx \= \kill
  \verb"input"\\
  \> $pageURLs\leftarrow$ URLs of pages to be stored\\
  \verb"output"\\
  \> $state\leftarrow$ the classifier output vectors for pages of
  $pageURLs$\\
  \verb"method"\\
  \> \verb"for each" $URL\,\in\,pageURLs$\\
  \> \> $page\leftarrow$ text of page of $URL$\\
  \> \> $state(URL)\leftarrow$ classifier output vector for $page$\\
  \> \verb"endfor"
  \end{tabbing}
  \hrule \vskip1pt \hrule
\end{pseudo}
\end{table}

A linear function approximator is used for LTP estimation. It encompasses $k$ parameters, the \textit{weight vector}
$\mathbf{weight}=(weight_1,\ldots , weight_k)$. The LTP of document of URL $A$ is estimated as the scalar product of
$\mathbf{state(A)}$ and $\mathbf{weight}$: $value(A)=\sum_{i=1}^k weight_i\,state(A)_i$. During URL ordering the URL
with highest LTP estimation is selected. The URL ordering algorithm is shown in Algorithm \ref{t:URLordering_pseudo}.

\begin{table}[htb]\hrule\vskip1pt\hrule\vskip2mm
\begin{pseudo}\label{t:URLordering_pseudo}\normalfont\textbf{URL Ordering}
  \vskip1mm \hrule %\vskip1pt \hrule
  \begin{tabbing}
  xxx \= xx \= xx \= xx \= xx \= xx \= xx \= xx \= xx \= \kill
  \verb"input"\\
  \> $frontier\leftarrow$ the set of available URLs\\
  \> $state\leftarrow$ the stored vector representation of the
  URLs\\
  \verb"output"\\
  \> $bestURL\leftarrow$ URL with maximum LTP value\\
  \verb"method"\\
  \> \verb"for each" $URL\,\in\,frontier$\\
  \> \> $value(URL)\leftarrow\sum_{i=1}^k state(URL)_i\, weight_i$\\
  \> \verb"endfor"\\
  \> $bestURL\leftarrow$ URL with maximal LTP $value$
  \end{tabbing}
  \hrule \vskip1pt \hrule
\end{pseudo}
\end{table}

The weight vector of each forager is tuned by Temporal Difference Learning
\cite{sutton88learning,szita03kalman,schultz00multiple}. Let us denote the current URL by $URL_n$, the next URL to be
visited by $URL_{n+1}$, the output of the classifier for $URL_j$ by $\mathbf{state(URL_j)}$ and the estimated LTP of a
URL $URL_j$ by $value(URL_j) = \sum_{i=1}^k wegiht_i\,state(URL_j)_i$. Assume that leaving $URL_n$ to $URL_{n+1}$ the
immediate profit is $r_{n+1}$. Our estimation is perfect if $value(URL_n)=value(URL_{n+1})+r_{n+1}$. Future profits are
typically discounted in such estimations as $value(URL_n)=\gamma value(URL_{n+1})+r_{n+1}$, where $0 < \gamma < 1$. The
error of value estimation is

$$\delta(n,n+1) = r_{n+1} + \gamma value(URL_{n+1}) - value(URL_n).$$

\noindent We used throughout the simulations $\gamma =0.9$. For each step $URL_n \rightarrow URL_{n+1}$ the weights of
the value function were tuned to decrease the error of value estimation based on the received immediate profit
$r_{n+1}$. The $\delta(n,n+1)$ estimation error was used to correct the parameters. The $i^{th}$ component of the
weight vector, $weight_i$, was corrected by

 $$\Delta weight_i = \alpha \,\delta(n,n+1) \, state(URL_n)_i$$

\noindent with $\alpha=0.1$ and $i=1, \ldots , k$. These modified weights in a stationary environment would improve
value estimation (see, e.g, \cite{Sutton98Reinforcement} and references therein). The URL ordering update is given in
Algorithm \ref{t:URLordering_update_pseudo}.

\begin{table}[htb]\hrule\vskip1pt\hrule\vskip2mm
\begin{pseudo}\label{t:URLordering_update_pseudo}\normalfont\textbf{URL Ordering Update}
  \vskip1mm \hrule %\vskip1pt \hrule
  \begin{tabbing}
  xxx \= xx \= xx \= xx \= xx \= xx \= xx \= xx \= xx \= \kill
  \verb"input"\\
  \> $URL_{n+1}\leftarrow$ the step for which the reinforcement is
  received\\
  \> $URL_n\leftarrow$ the previous step before $URL_{n+1}$\\
  \> $r_{n+1}\leftarrow$ reinforcement for visiting $URL_{n+1}$\\
  \verb"output"\\
  \> $weight\leftarrow$ the updated weight vector\\
  \verb"method"\\
  \> $\delta(n,n+1) \leftarrow r_{n+1} + \gamma value(URL_{n+1}) -
  value(URL_{n})$\\
  \> $weight \leftarrow weight\,+\,\alpha \,\delta(n,n+1) \, state(URL_n)$
  \end{tabbing}
  \hrule \vskip1pt \hrule
\end{pseudo}
\end{table}

Without the update algorithm the weight vector remains the same throughout the search.

\subsubsection{Document relevancy}\label{sss:relevant}

A document or page is possibly relevant for a forager if it is not
older than 24 hours and the forager has not marked it previously.
Algorithm \ref{t:relevant_pseudo} shows the procedure of selecting
such documents. The selected documents are sent to the RA for
further evaluation.

\begin{table}[htb]\hrule\vskip1pt\hrule\vskip2mm
\begin{pseudo}\label{t:relevant_pseudo}\normalfont\textbf{Document Relevancy at a forager}
  \vskip1mm \hrule %\vskip1pt \hrule
  \begin{tabbing}
  xxx \= xx \= xx \= xx \= xx \= xx \= xx \= xx \= xx \= \kill
  \verb"input"\\
  \> $pages\leftarrow$ the pages to be examined\\
  \verb"output"\\
  \> $relevantPages\leftarrow$ the selected pages\\
  \verb"method"\\
  \> $previousPages\leftarrow$ previously selected relevant
  pages\\
  \> $relevantPages\leftarrow$ all pages from $pages$ which are\\
  \> \> not older than 24 hours and\\
  \> \> not contained in $previousPages$\\
  \> $previousPages\leftarrow$ add $relevantPages$ to $previousPages$
  \end{tabbing}
  \hrule \vskip1pt \hrule
\end{pseudo}
\end{table}

\subsubsection{Multiplication of a forager}\label{sss:multiplication}

During multiplication the weblog is randomly divided into two
equal sized parts (one for the original and one for the new
forager). The parameters of the URL ordering algorithm (the weight
vector of the value estimation) are either copied or new random
parameters are generated. If the forager has a URL ordering update
algorithm then the parameters are copied. If the forager does not
have any URL ordering update algorithm then new random parameters
are generated, as shown in Algorithm
\ref{t:multiplication_pseudo}.

\begin{table}[htb]\hrule\vskip1pt\hrule\vskip2mm
\begin{pseudo}\label{t:multiplication_pseudo}\normalfont\textbf{Multiplication}
  \vskip1mm \hrule %\vskip1pt \hrule
  \begin{tabbing}
  xxx \= xx \= xx \= xx \= xx \= xx \= xx \= xx \= xx \= \kill
  \verb"input"\\
  \> $weblog$ \\
  \> weight vector of URL ordering\\
  \verb"output"\\
  \> $newWeblog$\\
  \> $newWeight$\\
  \verb"method"\\
  \> $newWeblog\leftarrow WEBLOG\_SIZE/2$ randomly selected\\
  \> \> URLs and values from $weblog$\\
  \> $weblog\leftarrow$ delete $newWeblog$ from $weblog$\\
  \> \verb"if" forager has URL ordering update algorithm \\
  \> \> $newWeight\leftarrow$ copy the weight vector of URL
  ordering\\
  \> \verb"else"\\
  \> \> $newWeight\leftarrow$ generate a new random weight
  vector\\
  \> \verb"endif"
  \end{tabbing}
  \hrule \vskip1pt \hrule
\end{pseudo}
\end{table}

\subsection{Reinforcing agent}\label{ss:ra}

A reinforcing agent controls the "life" of foragers. It can start,
stop, multiply or delete foragers. RA receives the URLs of
documents selected by the foragers, and responds with
reinforcements for the received URLs. The response is $REWARD=100$
(a.u.) for a relevant document and $PENALTY=-1$ (a.u.) for a not
relevant document. A document is relevant if it is not yet seen by
the reinforcing agent and it is not older than 24 hours. The
reinforcing agent maintains the score of each forager working for
it. Initially each forager has $INIT\_SCORE=100$ score. When a
forager sends a URL to the RA, the forager's score is decreased by
$SCORE-=0.05$. After each relevant page sent by the forager, the
forager's score is increased by $SCORE+=1$ (Algorithm
\ref{t:manageURL_pseudo}).

\begin{table}[htb]\hrule\vskip1pt\hrule\vskip2mm
\begin{pseudo}\label{t:manageURL_pseudo}\normalfont\textbf{Manage Received URL}
\vskip1mm\hrule %\vskip1pt \hrule
\begin{tabbing}
  xxx \= xx \= xx \= xx \= xx \= xx \= xx \= xxxxx \= \kill
  \verb"input"\\
  \> $URL,forager\leftarrow$ received URL from forager\\
  \verb"output"\\
  \> reinforcement to forager\\
  \> updated forager score\\
  \verb"method"\\
  \> $relevants\leftarrow$ relevant pages seen by the RA\\
  \> $page\leftarrow$ get page of $URL$ \\
  \> decrease forager's score with $SCORE-$\\
  \> \verb"if" $page\in relevants$ or page date is older than 24 hours\\
  \> \> send $PENALTY$ to forager\\
  \> \verb"else"\\
  \> \> $relevants\leftarrow$ add $page$ to $relevants$\\
  \> \> send $REWARD$ to forager\\
  \> \> increase forager's score with $SCORE+$\\
  \> \verb"endif"
\end{tabbing}
\hrule \vskip1pt \hrule
\end{pseudo}
\end{table}

When the forager's score reaches $MAX\_SCORE=200$ and the number
of foragers is smaller than $MAX\_FORAGER=16$ then the forager is
multiplied. That is a new forager is created with the same
algorithms as the original one has, but with slightly different
parameters. When the forager's score goes below $MIN\_SCORE=0$ and
the number of foragers is larger than $MIN\_FORAGER=2$ then the
forager is deleted (Algorithm \ref{t:manageForager_pseudo}). Note
that a forager can be multiplied or deleted immediately after it
has been stopped by the RA and before the next forager is
activated.

\begin{table}[htb]\hrule\vskip1pt\hrule\vskip2mm
\begin{pseudo}\label{t:manageForager_pseudo}\normalfont\textbf{: Manage
Forager} \vskip1mm \hrule %\vskip1pt \hrule
\begin{tabbing}
  xxx \= xx \= xx \= xx \= xx \= xx \= xx \= xxxxx \= \kill
  \verb"input"\\
  \> $forager\leftarrow$ the forager to be multiplied or deleted\\
  \verb"output"\\
  \> possibly modified list of foragers\\
  \verb"method"\\
  \> \verb"if" ($forager$'s score \,$\geq$\, $MAX\_SCORE$ \verb"and" \\
  \> \> \> number of foragers $<$ $MAX\_FORAGER$)\\
  \> \> $weblog,URLordering\leftarrow$ call $forager$'s \\
  \> \> \> \> \textbf{Multiplication, Alg. \ref{t:multiplication_pseudo}}\\
  \> \> \> $forager$ may modify it's own weblog\\
  \> \> $newForager\leftarrow$ create a new forager with the received\\
  \> \> \> $weblog$ and $URLordering$ \\
  \> \> set the two foragers' score to $INIT\_SCORE$\\
  \> \verb"else if" ($forager$'s score \,$\leq$\, $MIN\_SCORE$
  \verb"and" \\
  \> \> \> number of foragers $>$ $MIN\_FORAGER$) \\
  \> \> delete $forager$\\
  \> \verb"endif"
\end{tabbing}
\hrule \vskip1pt \hrule
\end{pseudo}
\end{table}

Foragers on the same computer are working in time slices one after
each other. Each forager works for some amount of time determined
by the RA. Then the RA stops that forager and starts the next one
selected by the RA. The pseudo-code of the reinforcing agent is
given in Algorithm \ref{t:reinforcing_pseudo}.

\begin{table}[htb]\hrule\vskip1pt\hrule\vskip2mm
\begin{pseudo}\label{t:reinforcing_pseudo}\normalfont\textbf{: Reinforcing Agent}
\vskip1mm \hrule %\vskip1pt \hrule
\begin{tabbing}
  xxx \= xx \= xx \= xx \= xx \= xx \= xx \= xxxxx \= \kill
  \verb"input"\\
  \> seed URLs\\
  \verb"output"\\
  \> $relevants\leftarrow$ found relevant documents\\
  \verb"method"\\
  \> $relevants\leftarrow$ empty set /*set of all observed relevant pages\\
  \> initialize $MIN\_FORAGER$ foragers with the seed URLs\\
  \> \> set one of them to be the next \\
  \> \verb"repeat" \\
  \> \> start next forager \\
  \> \> \> receive possibly relevant URL \\
  \> \> \> \> call \textbf{Manage Received URL, Alg. \ref{t:manageURL_pseudo}} with URL\\
  \> \> stop forager if its time period is over \\
  \> \> call \textbf{Manage Forager, Alg. \ref{t:manageForager_pseudo}} with this forager \\
  \> \> choose next forager \\
  \> \verb"until" time is over
\end{tabbing}
\hrule \vskip1pt \hrule
\end{pseudo}
\end{table}

\subsection{Foragers}\label{ss:forager}

A forager is initialized with parameters defining the URL
ordering, and either with a weblog or with a seed of URLs
(Algorithm \ref{t:initforager_pseudo}). After its initialization a
forager crawls in search paths, that is after a given number of
steps the search restarts and the steps between two restarts form
a path. During each path the forager takes $MAX\_STEP=100$ number
of steps, i.e., selects the next URL to be visited with a URL
ordering algorithm. At the beginning of a path a URL is selected
randomly  from the starting URL list. This list is formed from the
10 first URLs of the weblog. The weblog contains the possibly good
starting URLs with their associated weblog values in descending
order. The weblog algorithm modifies the weblog and so thus the
starting URL list before a new path is started. When a forager is
restarted by the RA, after the RA has stopped it, the forager
continues from the internal state in which it was stopped. The
pseudo code of step selection is given in Algorithm
\ref{t:stepselection_pseudo}.

\begin{table}[htb]\hrule\vskip1pt\hrule\vskip2mm
\begin{pseudo}\label{t:initforager_pseudo}\normalfont\textbf{Initialization of the forager}
  \vskip1mm \hrule %\vskip1pt \hrule
  \begin{tabbing}
  xxx \= xx \= xx \= xx \= xx \= xx \= xx \= xx \= xx \= \kill
  \verb"input"\\
  \> weblog or seed URLs \\
  \> URL ordering parameters \\
  \verb"output"\\
  \> initialized forager\\
  \verb"method"\\
  \> set path step number to $MAX\_STEP+1$  /*start new path\\
  \> set the weblog\\
  \> \> either with the input weblog \\
  \> \> or put the seed URLs into the weblog with 0 weblog value \\
  \> set the URL ordering parameters in URL ordering algorithm
  \end{tabbing}
  \hrule \vskip1pt \hrule
\end{pseudo}
\end{table}

\begin{table}[htb]\hrule\vskip1pt\hrule\vskip2mm
\begin{pseudo}\label{t:stepselection_pseudo}\normalfont\textbf{URL Selection}
  \vskip1mm \hrule %\vskip1pt \hrule
  \begin{tabbing}
  xxx \= xx \= xx \= xx \= xx \= xx \= xx \= xx \= xx \= \kill
  \verb"input"\\
  \> $frontier\leftarrow$ set of URLs available in this step\\
  \> $visited\leftarrow$  set of visited URLs in this path\\
  \verb"output"\\
  \> $step\leftarrow$ selected URL to be visited next\\
  \verb"method"\\
  \> \verb"if" path step number $\leq MAX\_STEP$\\
  \> \> $step\leftarrow$ selected URL by \textbf{URL Ordering, Alg. \ref{t:URLordering_pseudo}}\\
  \> \> increase path step number\\
  \> \verb"else"\\
  \> \> call the \textbf{Weblog Update, Alg. \ref{t:weblog_pseudo}} to update the weblog\\
  \> \> $step\leftarrow$ select a random URL from the starting URL
  list\\
  \> \> set path step number to 1\\
  \> \> $frontier\leftarrow$ empty set\\
  \> \> $visited\leftarrow$ empty set\\
  \> \verb"endif"
  \end{tabbing}
  \hrule \vskip1pt \hrule
\end{pseudo}
\end{table}

The URL ordering algorithm selects a URL to be the next step from
the frontier URL set. The selected URL is removed from the
frontier and added to the visited URL set to avoid loops. After
downloading the pages, only those URLs (linked from the visited
URL) are added to the frontier which are not in the visited set.

In each step the forager downloads the page of the selected URL
and all of the pages linked from the page of selected URL. It
sends the URLs of the possibly relevant pages to the reinforcing
agent. The forager receives reinforcements on any previously sent
but not yet reinforced URLs and calls the URL ordering update
algorithm with the received reinforcements. The pseudo code of a
forager is shown in Algorithm \ref{t:forager_pseudo}.

\begin{table}[htb]\hrule\vskip1pt\hrule\vskip2mm
\begin{pseudo}\label{t:forager_pseudo}\normalfont\textbf{Forager}
  \vskip1mm \hrule %\vskip1pt \hrule
  \begin{tabbing}
  xxx \= xx \= xx \= xx \= xx \= xx \= xx \= xx \= xx \= \kill
  \verb"input"\\
  \> $frontier\leftarrow$ set of URLs available in the next step\\
  \> $visited\leftarrow$  set of visited URLs in the current path\\
  \verb"output"\\
  \> sent documents to the RA\\
  \> modified $frontier$ and $visited$\\
  \> modified $weblog$ and URL ordering weight vector\\
  \verb"method"\\
  \> \verb"repeat" \\
  \> \> $step\leftarrow$ call \textbf{URL Selection, Alg. \ref{t:stepselection_pseudo}}\\
  \> \> $frontier\leftarrow$ remove $step$ from $frontier$\\
  \> \> $visited\leftarrow$ add $step$ to $visited$\\
  \> \> $page\leftarrow$ download the page of $step$\\
  \> \> $linkedURLs\leftarrow$ links of $page$\\
  \> \> $newURLs\leftarrow$ $linkedURLs$ which are not $visited$\\
  \> \> $frontier\leftarrow$ add $newURLs$ to $frontier$\\
  \> \> download pages of $linkedURLs$ \\
  \> \> call \textbf{Page Information Storage, Alg. \ref{t:pageinfo_URLordering_pseudo}} with $newURLs$\\
  \> \> $relevantPages\leftarrow$ call \textbf{Document Relevancy, Alg.
  \ref{t:relevant_pseudo}} for\\
  \> \> \> all pages \\
  \> \> send $relevantPages$ to reinforcing agent \\
  \> \> receive reinforcements for sent but not yet reinforced pages
  \\
  \> \> call \textbf{URL Ordering Update, Alg. \ref{t:URLordering_update_pseudo}} with\\
  \> \> \> the received reinforcements \\

  \> \verb"until" time is over
  \end{tabbing}
  \hrule \vskip1pt \hrule
\end{pseudo}
\end{table}

\clearpage

\section{Experiments}\label{s:experiments}

We conducted an 18 day long experiment on the Web to gather
realistic data. We used the gathered data in simulations to
compare the weblog update (Section \ref{sss:weblog}) and
reinforcement learning algorithms (Section \ref{sss:rl}). In Web
experiment we used a fleet of foragers using combination of
reinforcement learning and weblog update algorithms to eliminate
any biases on the gathered data. First we describe the experiment
on the Web then the simulations. We analyze our results at the end
of this section.

\subsection{Web}\label{ss:real}

We ran the experiment on the Web on a single personal computer
with Celeron 1000 MHz processor and 512 MB RAM. We implemented the
forager architecture (described in Section \ref{s:architecture})
in Java programming language.

In this experiment a fixed number of foragers were competing with each other to collect news at the CNN web site. The
foragers were running in equal time intervals in a predefined order. Each forager had a 3 minute time interval and
after that interval the forager was allowed to finish the step started before the end of the time interval. We deployed
8 foragers using the weblog update and the reinforcement learning based URL ordering update algorithms (8 WLRL
foragers). We also deployed 8 other foragers using the weblog update algorithm but without reinforcement learning (8 WL
foragers). The predefined order of foragers was the following: 8 WLRL foragers were followed by the 8 WL foragers.

\begin{figure}[htb]
  \centering
  \includegraphics[width=2.5in]{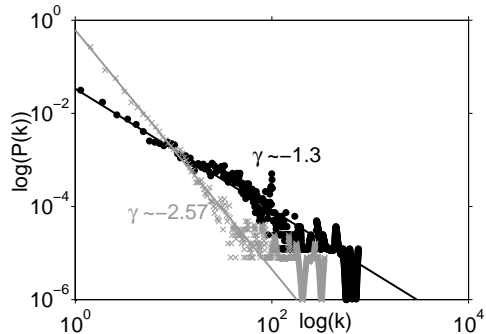}
  \caption{\textbf{Scale-free property of the Internet domain}. Log-log scale distribution of
the number of (incoming and outgoing) links of all URLs found
during the time course of investigation.  Horizontal axis: number
of edges ($\log k$). Vertical axis: relative frequency of number
of edges at different URLs ($\log P(k)$). Dots and dark line
correspond to outgoing links, crosses and gray line correspond to
incoming links.
  }\label{f:sf}
\end{figure}

We investigated the link structure of the gathered Web pages. As it
is shown in Fig. \ref{f:sf} the links have a power-law distribution
($P(k)=k^\gamma$) with $\gamma = -1.3$ for outgoing links and
$\gamma = -2.57$ for incoming links. That is the link structure has
the scale-free property. The clustering coefficient
\cite{watts98collective} of the link structure is 0.02 and the
diameter of the graph is 7.2893. We applied two different random
permutations to the origin and to the endpoint of the links, keeping
the edge distribution unchanged but randomly rewiring the links. The
new graph has 0.003 clustering coefficient and 8.2163 diameter. That
is the clustering coefficient is smaller than the original value by
an order of magnitude, but the diameter is almost the same.
Therefore we can conclude that the links of gathered pages form
small world structure.

The data storage for simulation is a centralized component. The pages are stored with 2 indices (and time stamps). One
index is the URL index, the other is the page index. Multiple pages can have the same URL index if they were downloaded
from the same URL. The page index uniquely identifies a page content and the URL from where the page was download. At
each page download of any foragers we stored the followings (with a time stamp containing the time of page download):

\begin{enumerate}
  \item if the page is relevant according to the RA then store
  ``relevant''
  \item if the page is from a new URL then store the new URL with a
  new URL index and the page's state vector with a new page index
  \item if the content of the page is changed since the last download then
  store the page's state vector with a new page index but keep the URL index
  \item in both previous cases store the links of the page as
  links to page indices of the linked pages
  \begin{enumerate}
    \item if a linked page is from a new URL then store the new URL
    with a new URL index and the linked page's state vector with a new page index
    \item if the content of the linked page is changed since the
    last check then store the page's state vector with a new page index but same
    URL index
  \end{enumerate}
\end{enumerate}

\subsection{Simulation}\label{ss:simulation}

For the simulations we implemented the forager architecture in
Matlab. The foragers were simulated as if they were running on one
computer as described in the previous section.

\subsubsection{Simulation specification}

During simulations we used the Web pages that we gathered
previously to generate a realistic environment (note that the
links of pages point to local pages (not to pages on the Web)
since a link was stored as a link to a local page index):
\begin{itemize}
  \item Simulated documents had the same state vector representation for URL
  ordering as the real pages had
  \item Simulated relevant documents were the same as the relevant
  documents on the Web
  \item Pages and links appeared at the same (relative) time when they were
  found in the Web experiment - using the new URL indices and their time stamps
  \item Pages and links are refreshed or changed at the same
  relative time as the changes were detected in the Web experiment -- using
  the new page indices for existing URL indices and their time
  stamps
  \item Simulated time of a page download was the average download
  time of a real page during the Web experiment.
\end{itemize}

We conducted simulations with two different kinds of foragers. The first case is when foragers used only the weblog
update algorithm without URL ordering update (WL foragers). The second case is when foragers used only the
reinforcement learning based URL ordering update algorithm without the weblog update algorithm (RL foragers). Each WL
forager had a different weight vector for URL value estimation -- during multiplication the new forager got a new
random weight vector. RL foragers had the same weblog with the first 10 URLs of the gathered pages -- that is the
starting URL of the Web experiment and the first 9 visited URLs during that experiment. In both cases initially there
were 2 foragers and they were allowed to multiply until reaching the population of 16 foragers. The simulation for each
type of foragers were repeated 3 times with different initial weight vectors for each forager. The variance of the
results show that there is only a small difference between simulations using the same kind of foragers, even if the
foragers were started with different random weight vectors in each simulation.

\subsubsection{Simulation measurements}

Table \ref{t:params} shows the investigated parameters during simulations.

\begin{table*}
\caption{Investigated parameters}\label{t:params} \centering
\begin{tabular}{|p{3cm}|p{10cm}|}\hline
  downloaded & is the number of downloaded documents \\
  sent & is the number of documents sent to the RA \\
  relevant & is the number of found relevant documents \\
  found URLs & is the number of found URLs \\
  download efficiency & is the ratio of relevant to downloaded documents in 3 hour time window
  throughout the simulation. \\
  sent efficiency & is the ratio of relevant to sent documents in 3 hour time window
  throughout the simulation. \\
  relative found URL & ratio of found URLs to downloaded at the end of the simulation \\
  freshness & is the ratio of the number of current found relevant documents and the number of all found relevant
  documents \cite{cho03effective}. A stored document is current, up-to-date, if its content is exactly the
  same as the content of the corresponding URL in the environment. \\
  age & A stored current document has 0 age, the age of an obsolete page is the time since
  the last refresh of the page on the Web \cite{cho03effective}.
  \\ \hline
\end{tabular}
\end{table*}

Parameter `download efficiency' is relevant for the site where the foragers should be deployed to gather the new
information while parameter `sent efficiency' is relevant for the RA. Note that during simulations we are able to
immediately and precisely calculate freshness and age values. In a real Web experiment it is impossible to calculate
these values precisely, because of the time needed to download and compare the contents of all of the real Web pages to
the stored ones.

\subsubsection{Simulation analysis}

The values in Table \ref{t:data} are averaged over the 3 runs of
each type of foragers.

\begin{table}[htb]
\caption{\textbf{Simulation results}. The $3^{rd}$ and $5^{th}$
columns contain the standard deviation of the individual
experiment results from the average values. }\label{t:data}
  \centering
  \vskip1mm
\begin{tabular}{|l|c|c|c|c|}\hline
  type & RL & std RL & WL & std WL\\ \hline
  downloaded            & 540636    & 9840      & 669673    & 9580  \\
  sent                  & 9747      & 98        & 6345      & 385   \\
  relevant              & 2419      & 45        & 3107      & 60    \\
  found URLs            & 31092     & 1050      & 33116     & 3370  \\
  download efficiency   & 0.0045    & 0.0001    & 0.0046    & 0.0001\\
  sent efficiency       & 0.248     & 0.003     & 0.49      & 0.031 \\
  relative found URL    & 0.058     & 0.001     & 0.05      & 0.006 \\
  freshness             & 0.7       & 0.006     & 0.74      & 0.011 \\
  age (in hours)        & 1.79      & 0.04      & 1.56      & 0.08  \\ \hline
\end{tabular}
\end{table}

From Table \ref{t:data} we can conclude the followings:
\begin{itemize}
  \item RL and WL foragers have similar download efficiency, i.e., the
  efficiencies from the point of view of the news site are about the same.
  \item WL foragers have higher sent efficiencies than RL foragers, i.e.,
  the efficiency from the point of view of the RA is higher. This shows that
  WL foragers divide the search area better among each other than
  RL foragers. Sent efficiency would be 1 if none of two foragers have sent the same document to the RA.
  \item RL foragers have higher relative found URL value than WL foragers.
  RL foragers explore more than WL foragers and RL found more
  URLs than WL foragers did per downloaded page.
  \item WL foragers find faster the new relevant documents in
  the already found clusters. That is freshness is higher and
  age is lower than in the case of RL foragers.
\end{itemize}

\begin{figure}[htb]
  \centering
  \includegraphics[width=2.5in]{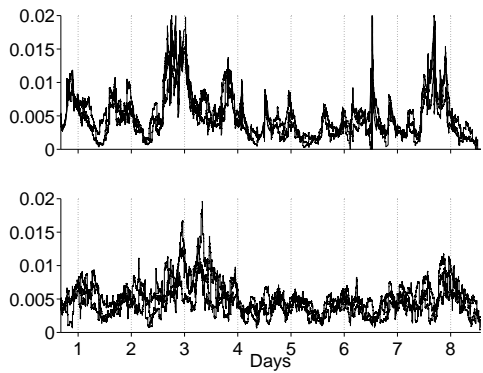}
  \caption{\textbf{Efficiency}. Horizontal axis: time in days.
  Vertical axis: download efficiency, that is the number of found relevant documents divided
  by number of downloaded documents in 3 hour time intervals.
  Upper figure shows RL foragers' efficiencies, lower figure
  shows WL foragers' efficiencies. For all of the 3 simulation experiments there
  is a separate line.
  }\label{f:efficiency}
\end{figure}

Fig. \ref{f:efficiency} shows other aspects of the different
behaviors of RL and WL foragers. Download efficiency of RL
foragers has more, higher, and sharper peaks than the download
efficiency of WL foragers has. That is WL foragers are more
balanced in finding new relevant documents than RL foragers. The
reason is that while the WL foragers remain in the found good
clusters, the RL foragers continuously explore the new promising
territories. The sharp peaks in the efficiency show that RL
foragers \textit{find and recognize} new good territories and then
\textit{quickly collect} the current relevant documents from
there. The foragers can recognize these places by receiving more
rewards from the RA if they send URLs from these places.

\begin{figure}[htb]
  \centering
  \includegraphics[width=2.5in]{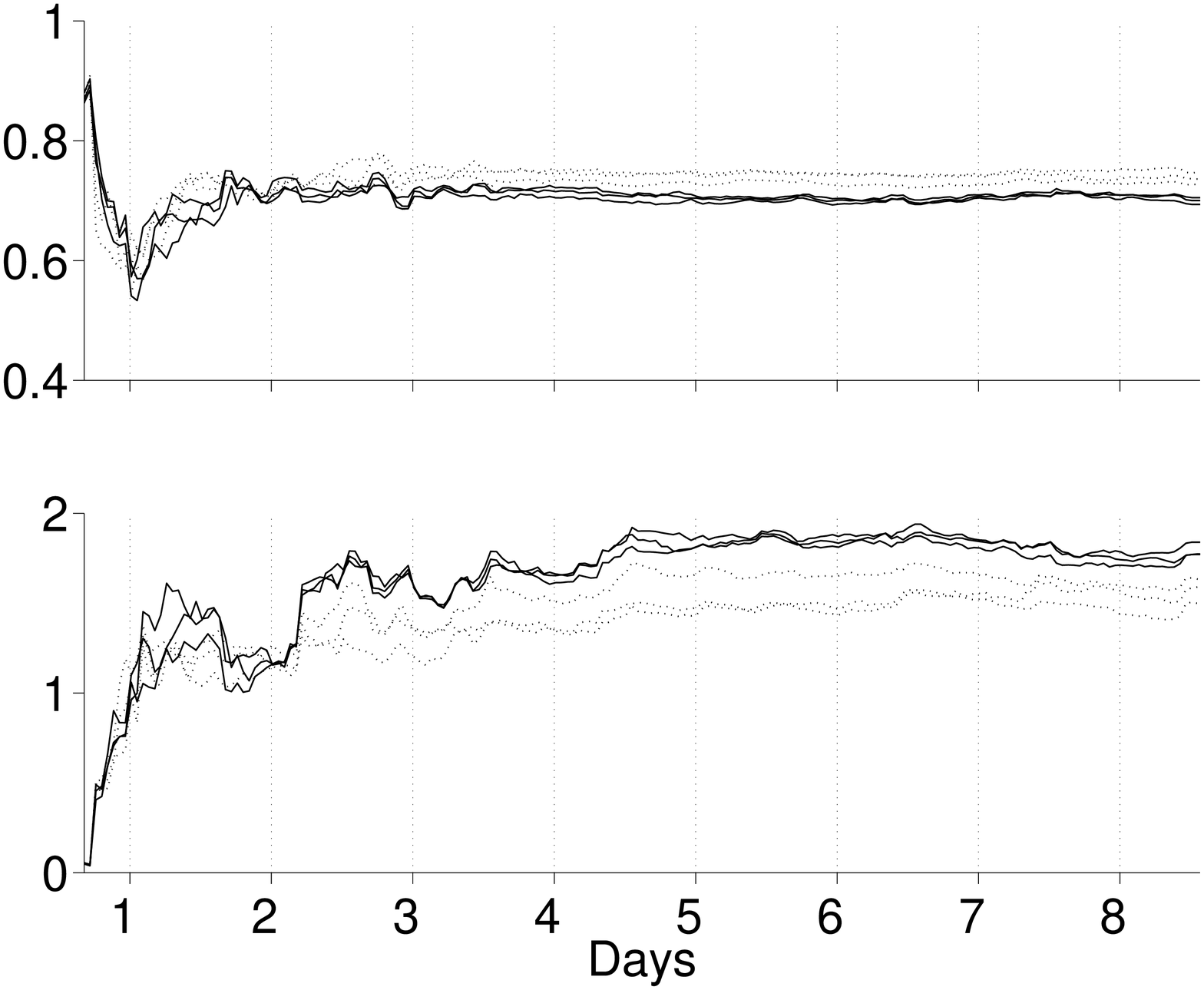}
  \caption{\textbf{Freshness and Age}. Horizontal axis: time in days.
  Upper vertical axis: freshness of found relevant documents in 3 hour time intervals.
  Lower vertical axis: age in hours of found relevant documents in 3 hour time intervals.
  Dotted lines correspond to weblog foragers, continuous lines correspond to RL foragers.
  }\label{f:freshness}
\end{figure}

The predefined order did not influence the working of foragers
during the Web experiment. From Fig. \ref{f:efficiency} it can be
seen that foragers during the 3 independent experiments did not
have very different efficiencies. On Fig. \ref{f:freshness} we
show that the foragers in each run had a very similar behavior in
terms of age and freshness, that is the values remains close to
each other throughout the experiments. Also the results for
individual runs were close to the average values in Table
\ref{t:data} (see the standard deviations). In each individual run
the foragers were started with different weight vectors, but they
reached similar efficiencies and behavior. This means that the
initial conditions of the foragers did not influence the later
behavior of them during the simulations. Furthermore foragers
could not change their environment drastically (in terms of the
found relevant documents) during a single 3 minute run time
because of the short run time intervals and the fast change of
environment -- large number of new pages and often updated pages
in the new site. During the Web experiment foragers were running
in 8 WLRL, 8 WL, 8 WLRL, 8 WL, \ldots temporal order. Because of
the fact that initial conditions does not influence the long term
performance of foragers and the fact that the foragers can not
change their environment fully we can start to examine them after
the first run of WLRL foragers. Then we got the other extreme
order of foragers, that is the 8 WL, 8 WLRL, 8 WL, 8 WLRL, \ldots
temporal ordering. For the overall efficiency and behavior of
foragers it did not really matter if WLRL or WL foragers run first
and one could use mixed order in which after a WLRL forager a WL
forager runs and after a WL forager a WLRL forager comes. However,
for higher bandwidths and for faster computers, random ordering
may be needed for such comparisons.

\section{Discussion}\label{s:discussion}

Our first conjecture is that selection is efficient on scale-free small world structures. L{\H o}rincz and K\'okai
\cite{Lorincz02intelligent} and Rennie et al. \cite{rennie99using} showed that RL is efficient in the task of finding
relevant information on the Web. Here we have shown experimentally that the weblog update algorithm, selection among
starting URLs, is at least as efficient as the RL algorithm. The weblog update algorithm finds as many relevant
documents as RL does if they download the same amount of pages. WL foragers in their fleet select more different URLs
to send to the RA than RL foragers do in their fleet, therefore there are more relevant documents among those selected
by WL foragers then among those selected by RL foragers. Also the freshness and age of found relevant documents are
better for WL foragers than for RL foragers.

For the weblog update algorithm, the selection among starting URLs
has no fine tuning mechanism. Throughout its life a forager
searches for the same kind of documents -- goes into the same
`direction' in the state space of document states -- determined by
its fixed weight vector. The only adaptation allowed for a WL
forager is to select starting URLs from the already seen URLs. The
WL forager can not modify its (`directional') preferences
according goes newly found relevant document supply, where
relevant documents are abundant. But a WL forager finds good
relevant document sources in its own direction and forces its
search to stay at those places. By chance the forager can find
better sources in its own direction if the search path from a
starting URL is long enough. On Fig. \ref{f:efficiency} it is
shown that the download efficiency of the foragers does not
decrease with the multiplication of the foragers. Therefore the
new foragers must found new and good relevant document sources
quickly after their appearances.

The reinforcement learning based URL ordering update algorithm is capable to fine tune the search of a forager by
adapting the forager's weight vector. This feature has been shown to be crucial to adapt crawling in novel environments
\cite{kokai02learning,Lorincz02intelligent}. An RL forager goes into the direction (in the state space of document
states) where the estimated long term cumulated profit is the highest. Because the local environment of the foragers
may changes rapidly during crawling, it seems desirable that foragers can quickly adapt to the found new relevant
documents. Relevant documents may appear lonely, not creating a good relevant document source, or do not appear at the
right URL by a mistake. This noise of the Web can derail the RL foragers from good regions. The forager may ``turn''
into less valuable directions, because of the fast adaptation capabilities of RL foragers.

Our second conjecture is that selection fits SFSW better than RL. We have shown in our experiments that selection and
RL have different behaviors. Selection selects good information sources, which are worth to revisit, and stays at those
sources as long as better sources are not found by chance. RL explores new territories, and adapts to those. This
adaptation can be a disadvantage when compared with the more rigid selection algorithm, which sticks to good places
until `provably' better places are discovered. Therefore WL foragers, which can not be derailed and stay in their found
`niches' can find new relevant documents faster in such already known terrains than RL foragers can. That is, freshness
is higher and age is lower for relevant documents found by WL foragers than for relevant documents found by RL
foragers. Also, by finding good sources and staying there, WL foragers divide the search task better than RL foragers
do, this is the reason for the higher sent efficiency of WL foragers than of RL foragers.

We have rewired the network as it was described in Section \ref{ss:real}. This way a scale-free (SF) but not so small
world was created. Intriguingly, in this SF structure, RL foragers performed better than WL ones. Clearly, further work
is needed to compare the behavior of the selective and the reinforcement learning algorithms in other then SFSW
environments. Such findings should be of relevance in the deployment of machine learning methods in different problem
domains.

From the practical point of view, we note that it is an easy matter to combine the present algorithm with URLs offered
by search engines. Also, the values reported by the crawlers about certain environments, e.g., the environment of the
URL offered by search engines represent the neighborhood of that URL and can serve adaptive filtering. This procedure
is, indeed, promising to guide individual searches as it has been shown elsewhere \cite{palotai04adaptive}.

\section{Conclusion}\label{s:conclusion}

We presented and compared our selection algorithm  to the
well-known reinforcement learning algorithm. Our comparison was
based on finding new relevant documents on the Web, that is in a
dynamic scale-free small world environment. We have found that the
weblog update selection algorithm performs better in this
environment than the reinforcement learning algorithm, eventhough
the reinforcement learning algorithm has been shown to be
efficient in finding relevant information
\cite{Lorincz02intelligent,rennie99using}. We explain our results
based on the different behaviors of the algorithms. That is the
weblog update algorithm finds the good relevant document sources
and remains at these regions until better places are found by
chance. Individuals using this selection algorithm are able to
quickly collect the new relevant documents from the already known
places because they monitor these places continuously. The
reinforcement learning algorithm explores new territories for
relevant documents and if it finds a good place then it collects
the existing relevant documents from there. The continuous
exploration and the fine tuning property of RL causes that RL
finds relevant documents slower than the weblog update algorithm.

In our future work we will study the combination of the weblog
update and the RL algorithms. This combination uses the WL
foragers ability to stay at good regions with the RL foragers fine
tuning capability. In this way foragers will be able to go to new
sources with the RL algorithm and monitor the already found good
regions with the weblog update algorithm.

We will also study the foragers in a simulated environment which
is not a small world. The clusters of small world environment
makes it easier for WL foragers to stay at good regions. The small
diameter due to the long distance links of small world environment
makes it easier for RL foragers to explore different regions. This
work will measure the extent at which the different foragers rely
on the small world property of their environment.

\section{Acknowledgement}

This material is based upon work supported by the European Office
of Aerospace Research and Development, Air Force Office of
Scientific Research, Air Force Research Laboratory, under Contract
No. FA8655-03-1-3036. This work is also supported by the National
Science Foundation under grants No. INT-0304904 and No.
IIS-0237782. Any opinions, findings and conclusions or
recommendations expressed in this material are those of the
author(s) and do not necessarily reflect the views of the European
Office of Aerospace Research and Development, Air Force Office of
Scientific Research, Air Force Research Laboratory.

\bibliographystyle{amsplain}

\begin{thebibliography}{10}

\bibitem{barabasi00scalefree}
A.L. Barab\'asi, R.~Albert, and H.~Jeong, \emph{Scale-free
characteristics of
  random networks: The topology of the world wide web}, Physica A \textbf{281}
  (2000), 69--77.

\bibitem{Boley98principal}
D.L. Boley, \emph{Principal direction division partitioning}, Data
Mining and
  Knowledge Discovery \textbf{2} (1998), 325--244.

\bibitem{cho03effective}
J.~Cho and H.~Garcia-Molina, \emph{Effective page refresh policies
for web
  crawlers}, ACM Transactions on Database Systems \textbf{28} (2003), no.~4,
  390--426.

\bibitem{Clarck00dynamic}
C.W. Clark and M.~Mangel, \emph{Dynamic state variable models in
ecology:
  \mbox{M}ethods and applications.}, Oxford University Press, Oxford UK, 2000.

\bibitem{Csanyi89evolutionary}
V.~Cs\'anyi, \emph{Evolutionary systems and society: \mbox{A}
general theory of
  life, mind, and culture}, Duke University Press, Durham, NC, 1989.

\bibitem{edwards01adaptive}
J.~Edwards, K.~McCurley, and J.~Tomlin, \emph{An adaptive model for
optimizing
  performance of an incremental web crawler}, Proceedings of the tenth
  international conference on World Wide Web, 2001, pp.~106--113.

\bibitem{eiben03introduction}
A.~E. Eiben and J.E. Smith, \emph{Introduction to evolutionary
computing},
  Springer, 2003.

\bibitem{Fryxell98individual}
J.M. Fryxell and P.~Lundberg, \emph{Individual behavior and
community
  dynamics.}, Chapman and Hall, London, 1998.

\bibitem{Gabor04value}
B.~G\'abor, Zs. Palotai, and A.~L{\H o}rincz, \emph{Value estimation
based
  computer-assisted data mining for surfing the internet}, Int. Joint Conf. on
  Neural Networks (Piscataway, NJ 08855-1331), IEEE Operations Center, 26-29
  July, Budapest, Hungary 2004, pp.~Paper No. 1035., IEEE Catalog Number:
  04CH37541C, IJCNN2004 CD--ROM Conference Proceedings.

\bibitem{joachims97probabilistic}
Thorsten Joachims, \emph{A probabilistic analysis of the {R}occhio
algorithm
  with {TFIDF} for text categorization}, Proceedings of {ICML}-97, 14th
  International Conference on Machine Learning (Nashville, US) (Douglas~H.
  Fisher, ed.), Morgan Kaufmann Publishers, San Francisco, US, 1997,
  pp.~143--151.

\bibitem{kampis91selfmodifying}
G.~Kampis, \emph{Self-modifying systems in biology and cognitive
science: A new
  framework for dynamics, information and complexity}, Pergamon, Oxford UK,
  1991.

\bibitem{Kleinberg01structure}
J.~Kleinberg and S.~Lawrence, \emph{The structure of the web},
Science
  \textbf{294} (2001), 1849--1850.

\bibitem{kokai02learning}
I.~K\'okai and A.~L{\H o}rincz, \emph{Fast adapting value estimation
based
  hybrid architecture for searching the world-wide web}, Applied Soft Computing
  \textbf{2} (2002), 11--23.

\bibitem{kondo04reinforcement}
T.~Kondo and K.~Ito, \emph{A reinforcement learning with
evolutionary state
  recruitment strategy for autonomous mobile robots control}, Robotics and
  Autonomous Systems \textbf{46} (2004), 11--124.

\bibitem{Lorincz02intelligent}
A.~L{\H o}rincz, I.~K\'okai, and A.~Meretei, \emph{Intelligent
high-performance
  crawlers used to reveal topic-specific structure of the \mbox{WWW}}, Int. J.
  Founds. Comp. Sci. \textbf{13} (2002), 477--495.

\bibitem{mataric97reinforcement}
Maja~J. Mataric, \emph{Reinforcement learning in the multi-robot
domain},
  Autonomous Robots \textbf{4} (1997), no.~1, 73--83.

\bibitem{menczer03complementing}
F.~Menczer, \emph{Complementing search engines with online web
mining agents},
  Decision Support Systems \textbf{35} (2003), 195--212.

\bibitem{moriarty99evolutionary}
D.E. Moriarty, A.C. Schultz, and J.J. Grefenstette,
\emph{Evolutionary
  algorithms for reinforcement learning}, Journal of Artificial Intelligence
  Research \textbf{11} (1999), 199--229.

\bibitem{taylor02mutualism}
E.~Pachepsky, T.~Taylor, and S.~Jones, \emph{Mutualism promotes
diversity and
  stability in a simple artificial ecosystem.}, Artificial Life \textbf{8}
  (2002), no.~1, 5--24.

\bibitem{palotai04adaptive}
Zs. Palotai, B.~G\'abor, and A.~L{\H o}rincz, \emph{Adaptive
highlighting of
  links to assist surfing on the internet}, Int. J. of Information Technology
  and Decision Making (2005), (to appear).

\bibitem{rennie99using}
J.~Rennie, K.~Nigam, and A.~McCallum, \emph{Using reinforcement
learning to
  spider the web efficiently}, Proc. 16th Int. Conf. on Machine Learning
  ({ICML}), Morgan Kaufmann, San Francisco, 1999, pp.~335--343.

\bibitem{risvik02search}
K.~M. Risvik and R.~Michelsen, \emph{Search engines and web
dynamics}, Computer
  Networks \textbf{32} (2002), 289--302.

\bibitem{rungsawang04learnable}
A.~Rungsawang and N.~Angkawattanawit, \emph{Learnable topic-specific
web
  crawler}, Computer Applications \textbf{xx} (2004), xxx--xxx.

\bibitem{schultz00multiple}
W.~Schultz, \emph{Multiple reward systems in the brain.}, Nature
Review of
  Neuroscience \textbf{1} (200), 199--207.

\bibitem{stafylopatis98autonomous}
A.~Stafylopatis and K.~Blekas, \emph{Autonomous vehicle navigation
using
  evolutionary reinforcement learning}, European Journal of Operational
  Research \textbf{108} (19998), 306--318.

\bibitem{sutton88learning}
R.~Sutton, \emph{Learning to predict by the method of temporal
differences},
  Machine Learning \textbf{3} (1988), 9--44.

\bibitem{Sutton98Reinforcement}
R.~Sutton and A.G. Barto, \emph{Reinforcement learning: An
introduction}, MIT
  Press, Cambridge, 1998.

\bibitem{szita03kalman}
I.~Szita and A.~L{\H o}rincz, \emph{Kalman filter control embedded
into the
  reinforcement learning framework.}, Neural Computation \textbf{16} (2004),
  491--499.

\bibitem{tesauro95temporal}
G.~J. Tesauro, \emph{Temporal difference learning and td-gammon},
Communication
  of the ACM \textbf{38} (1995), 58--68.

\bibitem{tuyls03extended}
K.~Tuyls, D.~Heytens, A.~Nowe, and B.~Manderick, \emph{Extended
replicator
  dynamics as a key to reinforcement learning in multi-agent systems}, ECML
  2003, LNAI 2837 (N.~Lavrac et~al., ed.), Springer-Verlag, Berlin, 2003,
  pp.~421--431.

\bibitem{watts98collective}
D.~J. Watts and S.~H. Strogatz, \emph{Collective dynamics of
`small-world'
  networks}, Nature \textbf{393} (1998), no.~6684, 440--442.

\bibitem{yao93review}
X.~Yao, \emph{Review of evolutionary artificial neural networks},
International
  Journal of Intelligent Systems \textbf{8} (1993), 539--567.

\bibitem{yao99evolving}
\bysame, \emph{Evolving artificial neural networks}, Proceedings of
the IEEE,
  vol.~87, 1999, pp.~1423--1447.

\end{thebibliography}
\providecommand{\bysame}{\leavevmode\hbox
to3em{\hrulefill}\thinspace}
\providecommand{\MR}{\relax\ifhmode\unskip\space\fi MR }
\providecommand{\MRhref}[2]{%
  \href{http://www.ams.org/mathscinet-getitem?mr=#1}{#2}
} \providecommand{\href}[2]{#2}

\end{document}